# Semantic Advertising

Ben Zamanzadeh[1], Naveen Ashish[2], Cartic Ramakrishnan[1] and John Zimmerman[1]
*DataPop Inc., 5762 W. Jefferson Blvd, Los Angeles 90016*
[2]*University of California Irvine, Irvine 92697*
*{ben, cartic, johnz}@datapop.com*

## 1. Introduction

This paper introduces the concept of online "Semantic Advertising", which we see as the technology that will help realize the full potential of Internet advertising. Internet advertising is a rapidly growing and arguably a dominant form of advertising. A recent IDC report (Weide, 2013) estimates that the total Internet advertising spend in 2011 was 87.4 billion dollars ($35B for the U.S. only), and predicts an annual growth rate of 16% over the next 5 years. We argue that Semantic Advertising, (SA), enables us to address the challenge of delivering relevance at scale in Internet Advertising.

Our argument is based on our work as a company developing *semantic technology* for better online advertising. Semantic technology (Hitzler, Krotzsch and Rudolph, 2009) can be described as algorithms and software that enable representation and reasoning based on meaning. Several companies such as Google, Microsoft and Yahoo, and smaller start-up companies have developed semantic technologies for advertising. The Advertising Technology (a.k.a *Ad Tech*) industry, including ad agencies, ad exchanges and campaign management tools has started to experiment with semantic technologies as well. The central argument put forth in this paper is that the *creation, analysis and management of effective online advertising campaigns at scale cannot be achieved without the use and continued evolution of semantic technologies and standards. Furthermore, Machine Intelligence cannot and should not replace human creativity in the creation and composition of ads.*

Traditional adverting is largely a manual creative process. Human intuition and experience play a large role in developing inventive and appealing messages for ads. Online advertising affords advertisers an unprecedented ability to simultaneously target highly granular consumer segments using individualized messaging strategies. However, this ability often results in a massive increase in data volume and complexity in online advertising. Managing the volume and complexity of data (a.k.a Big Data) therefore becomes the primary focus in the online-ad-campaign design process, thereby reducing a once creative process to one that involves the manual drudgery of data management. Semantic technology solves this issue by mapping the Big Data to a semantic space where ad writers can focus on marketing strategies and the system manages the individualization at scale.

This paper is organized as follows. In the next section i.e., Section 2 we place SA in broader semantic concepts related to SA such as the Semantic-Web, Semantic E Commerce, Semantic Marketing and others. In Section 3 we put forth the value of semantic technology in advertising. We discuss the limitations of the current state-of-the-art, and present our concept of the "ideal" ad which is what semantic advertising must strive towards. Section 4 provides a survey of current work in semantic advertising in areas such as context matching, semantic matching and optimization. Section 5 covers the semantic web technologies that have been proposed for and are relevant to semantic advertising. In Section 6 we emphasize the importance of domain knowledge and ontologies. Finally Section 7 touches on areas of future work and provides a conclusion.

## 2. Semantic Technology and Advertising

Semantic technology in general has been an area of active discussion, research, technology development and even controversy for over a decade – thus we wish to briefly cover the space and where SA fits in here. We place SA in the overall broader contexts of *Semantic Search*, *Semantic E-commerce*, *Semantic Marketing* and the vision of the *Semantic-Web*, as illustrated in Figure 1.

**The *Semantic-Web*** (Berners-Lee, Hendler and Lassila, 2001) has been described as "a collaborative movement led by the international standards body, the World Wide Web Consortium (W3C), that provides a common framework that allows data to be shared and reused across application, enterprise, and community boundaries".

**Semantic *Search*** (John, T. 2012; Grimes, S. 2010) is an information retrieval technology which seeks to improve search accuracy by understanding searcher intent and the contextual meaning of terms as they appear in the searchable space, whether on the Web or within a closed system, to generate more relevant results. Major web search engines like Google and Bing incorporate some elements of semantic search into their offerings, and Google's semantic search offering leverages Knowledge Graph technology – which we discuss later.

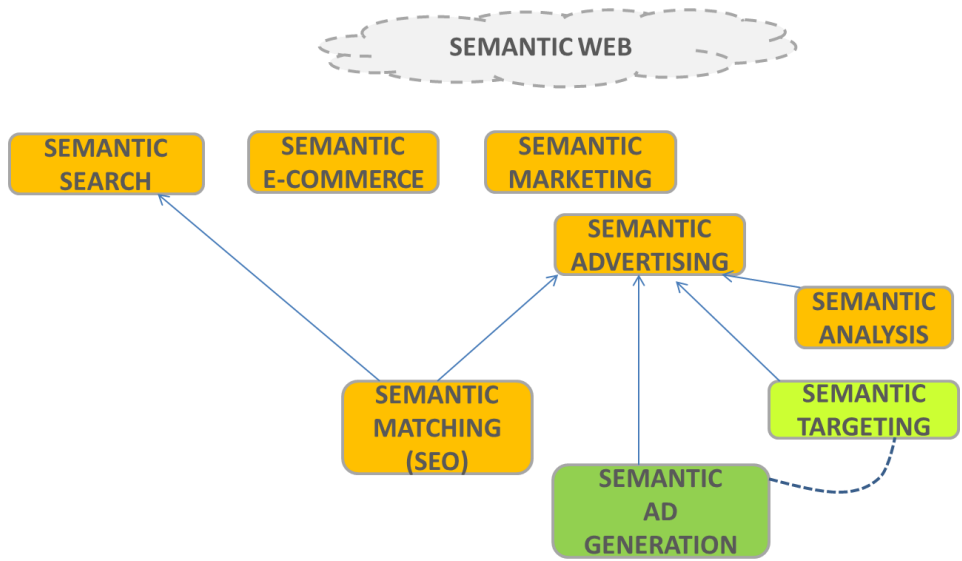

Figure 1 Semantic Advertising in Semantic E-Commerce

Semantic search enriches intelligent discovery of content relevant to a user on the Internet by using the explicit and inferred meaning of content based on the relationships in the semantic networks (i.e. Knowledge Graph).

**Semantic E** *commerce* is the intersection of semantic technology and online shopping. Semantic ecommerce uses semantic insights to direct the consumer within the ecommerce site and make specific offers to consumers. This technology uses insights learned from social sites and consumer reviews to improve the consumer conversion on the ecommerce site.

**Semantic *Marketing*** The term Semantic Marketing (Varadarajan and Yadav, 2011; Chen at al., 2011; Yannopoulos 2009) is an over-loaded term and means different things to different industries. The SEO industry thinks of it as using semantic tagging and annotation to enhance the web sites search ranking (related to Semantic Search). Social-Media thinks of it as generating *semantic tags* for consumer data to identify consumer intent and potential shopping tendencies. Semantic Marketing can also be considered, as the strategy that places paramount importance on the meaning and inherent message embedded in the ad based on both the ad copy and visuals. For the purposes of this paper we are focused on Semantic Advertising as it relates to ad copy or the precise selection of words and terms that convey an exact message which is the key to the advertising strategy.

**Semantic *Targeting*** is a technique used in selection and classification of a targeted audience based on the meaning and inferred semantic relationships in the content generated by the consumers (i.e. search queries, blogs, tweets, etc.,) or by publishers (Web pages, video content, etc.,). This capability enables the scaling of advertisements by automated systems based on the automated semantic targeting.

**Semantic *Analysis*** While the power of learning from large and complex data (Big Data) using machined learned algorithms has been a huge step, the inclusion of semantic data models and semantic reasoning to the mix has resulted in major breakthroughs in analyzing advertising performance data. It is semantics that sheds light on big data, which by itself is "dark" data, the semantic meaning, and relationships can convert this big dark data into *meaningful* data. Semantic Analysis enables modeling the problem in cognitive and abstract terms and we believe that that is critical to realizing a meaningful way of providing the right ad to the right consumer. Semantic Analysis is a key ingredient of Semantic e-Commerce, search, matching marketing and advertising.

**Semantic *Matching*** is the technology of providing relevant ads, in response to a search query, and discussed in more detail ahead.

## 3. Semantic Ad Generation

Unfortunately, the reality today is that online ads are often far from "satisfactory" or appealing and users are bombarded with misplaced and irrelevant ad messages. Some of the most frequent problems we see with online ads include: 1) A partial or even total irrelevance of the ad to the user intent, 2) "Robotic" tone of ad, and 3) A lack of value proposition, and key differentiators. Quite often human creativity is eroded from messages within ads reducing them to little more than factual 'monochromatic' descriptions of products. Although contextual matching is essential for selecting ads from the pool of available ads, the processes of generating the ads and decision-making required is massively improved by employing semantic technologies. What Semantic Advertising aims to do is to infuse meaningfulness with semantics into the online advertising process. Further evolution of Ad Tech and the appearance of Social, Mobile and Native Ads has led the way for appearance of "Individualized Ads" where successful ads are tailored to the individual's intent, needs, and characteristics (relate to consumer). Semantic Ad Generation makes it possible to employ specific marketing strategies at scale by incorporating individualization as an integral part of each ad .

Bad ads are easy enough to identify, but let us consider what makes a *good* ad. A notion of good ad is proposed in the "Conversion Trinity" developed by Bryan Eisenberg (Eisenberg, 2011) where he explains a three step formula comprising of Relevance, Value and Call-to-action.

- **Relevance.** Are you relevant to *my* wants/needs/desires (search query)?
- **Value.** Do I know *why* you are the right solution for me? Have you explained your value proposition/offer well?
- **Call to action.** Is it obvious *what* I need to do next? Have you given me the confidence to take that action?

## What Makes a Great Ad ?

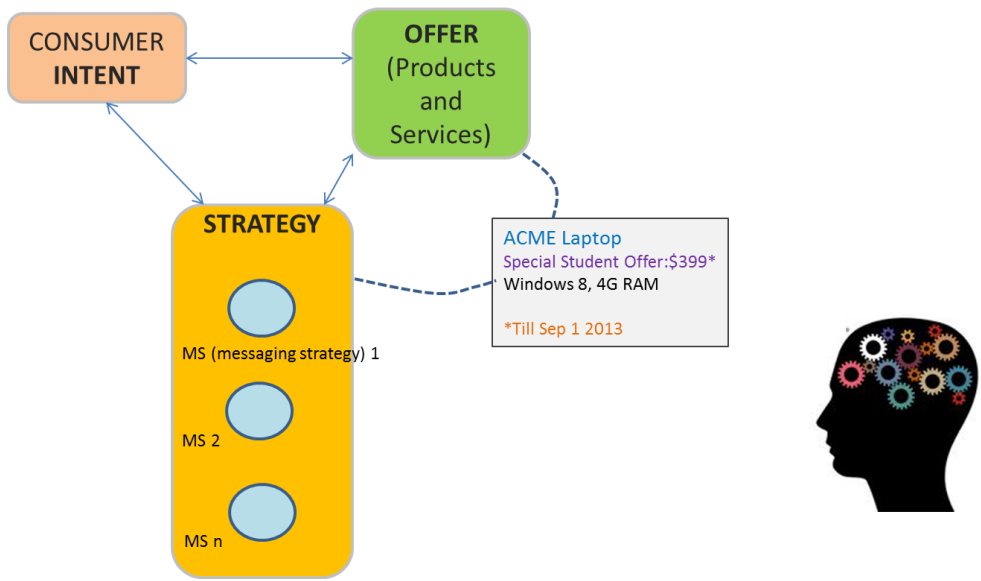

Figure 2 Relevant, Valuable and Actionable Advertisements

Generating relevant offers to individual characteristics of each consumer at internet scale has been the core issue of traditional advertising technologies. DataPop's experience in formulating a process to build effective advertising at a large scale has identified 3 components that every successful ad must contain as displayed in Figure 2. We have incorporated the "Conversion Trinity" into our machine learned optimization algorithms to build ads that optimize conversion along with satisfying the advertisers' key objectives and strategies. Our work has resulted in the evolution of semantic models representing ads, which in turn use semantic models for relevance, offer and advertising strategies. Our resulting trinity is:

1) **Relevance:** Ads must be relevant to the consumer's *intent and targeted audience*, including audience characteristics such as their demographics, etc. Relevance is actually represented by relationship vectors between an ad and consumer intent or targeted audience characteristics.

2) **Offer:** Ads must offer certain goods, services, solutions and/or a specific *value proposition* to the consumer.

3) **Advertising Strategy:** Various wordings and presentation methods used to convince the consumer to take action or establish certain consumer sentiment.

Semantic Advertising achieves meaningful and optimized advertising through the use of *semantic meta-data* as well as a *semantic network* (Hitzler, Krotzsch and Rudolph, 2009) of concepts and relationships linking ads to intent and people. We have developed a Marketing and Advertising Semantic Object Network (a.k.a. "MASON"), which entails representing real world advertising entities in form of *semantic object models* (Hammer and McLeod, 1990) and building a network of relationships amongst them. Advertising concepts such as consumer intent, target, offer, product, services, solution, and various advertising strategies are converted into semantic object models and then machine learning algorithms are used to generate advertising campaigns including campaign structure and creative.

Semantic Ad Generation uses semantic object models for advertising strategies established by the advertisers to develop *"Ad Creatives"*. Advertising strategies include a range of techniques and strategies used to generate the presentation, content and inherent message of the advertising. These strategies include Messaging strategies, Targeting strategies, Offering Strategies, Presentation Strategies, etc. Targeting Strategies are used to generate *semantic targets* so that the proper messaging strategy is applied to the proper target. The DataPop system builds semantic data models of advertising strategies using semantic encoding, inference and reasoning engines.This technology greatly reduces manual drudgery of data management while transforming the online-ad-campaign design process into a creative one driven by human intuition and enabling the simultaneous targeting highly granular consumer segments on a strategic level. The key aspect of DataPop's technology that makes this goal achievable is the use of semantic metadata, the use of which transforms the field of *Computational Advertising* - which is defined as "the field of algorithms that provides the best ad to an Internet user" (Broder, 2008), into Semantic Advertising.

# 4. Current Work

There has been a significant amount of work related to semantic advertising. Most of this work falls in the area of semantic (context) matching and the associated optimization issues in the ad network. In this section we provide a summary overview of work in these areas.

## Semantic Search

The term Semantic Search was coined in (Guha, McCool and Miller, 2003) defined as "the application of Semantic-Web to search". The key aim of Semantic Search was defined as and continues to be infusing meaning in search, as opposed to search technology based on just keyword matching. Semantic Search is key for SA as it is the technology that can enables a meaningful discovery phase i.e., semantically match user requests to products and relevant ads. In the last 5 years Semantic Search has become a technology underlying applications that we use daily, namely in 1) Google, with its acquisition and subsequent integration of "Freebase" (Freebase, 2013) – a semantic knowledge base and also the realization of Google's Knowledge Graph - a semantic network that drives Semantic Search, 2) Microsoft Bing, with its acquisition and integration of Semantic Search technology from "Powerset" – a Silicon valley Semantic Search start-up company based on natural language processing (NLP) technology, and 3) Facebook's "Graph Search" that is facilitating natural language semantic querying of information over the Facebook graph of concepts and relationships. Core Semantic Search technologies such as domain ontologies, natural language query understanding and semantic relevance matching are highly relevant to SA.

## Context Matching

Context Match(CM) is the area where we have seen the majority of the work in online advertising so far. For instance (Broder et al., 2007) propose a way of matching advertisements to Web pages, where the semantic match (between an ad and a page) is a major component of the final relevance score. Pages and ads are classified into a pre-assembled advertising ontology or, strictly speaking, a taxonomy. The semantic match is *combined* with a syntactic match and the final relevance score is a weighted combination of the syntactic and semantic scores. (Anagnostopoulos et al, 2011) build further upon the work in (Broder et al., 2007) for better real-time context matching of Web pages. Text summarization techniques are used to identify short but informative page fragments that can serve as a good proxy for the entire page. Further, sources of *external knowledge* are also employed.

A technique called "SIWI" employs Wikipedia based matching (Wu et al., 2012) to improve contextual match. Traditional keyword matching has the problems caused by homonymy and polysemy, low intersection of textual keywords and context mismatch. In the SIWI approach a select subset of Wikipedia articles are identified and used as an *intermediate semantic reference model* to match pages with ads. Wikipedia articles that are relevant to a page are automatically identified and a semantic representation of the page is created using the Wikipedia articles – which is in turn used to match ads with that page. (Phuong and Phuong, 2012) propose using topic modeling to improve the relevance of retrieved ads. This method uses a keyword-topic model that associates each keyword provided by the advertiser with a multinomial distribution over topics. Then, an ad with multiple keywords is represented as a mixture of topic distributions associated with those keywords. (Mirizzi et al., 2010) present a Semantic Tagging approach to contextual matching where they exploit semantic relations stored in the DBpedia dataset and use a hybrid ranking system to rank keywords and to expand queries formulated by the user. (Massimiliano and Vassilis, 2009) present an approach called semantic associations which is a machine learning approach to contextual advertising where a novel set of features is used to capture subtle semantic associations – frequently occurring collocated words, between the vocabularies of the ad and the Web page. A model for rankings ads is then automatically learnt. Other works include (Zhang et al., 2008) which actually identifies Web pages that an ad should *not* be place on, based on determining content that the advertiser would actually wish to avoid being associated with.

## 5. Semantic-Web Technologies and Framework

SA is based on semantic technologies which in turn have their roots in the Semantic-web vision itself. Over the years many technologies, frameworks and tools have been proposed and explored with varying success and it is important that we introduce the relevant technologies along with their degree of adoption and prevalence today.

### Semantic Technologies

The initial vision of the Semantic-Web was based on, amongst other things, a rich semantic "markup" of Web pages in *markup languages* such as *RDF* and *OWL*. RDF stands for the Resource Description Framework and OWL for Web Ontology Language (Hitzler, Krotzsch and Rudolph, 2009). These are XML based languages to markup the content of Web pages in semantic terms. They are also formats in which ontologies can be represented and reasoned with. More recent paradigms, especially those that are more "lightweight" have received considerably more traction than RDF itself. *Microdata* (Microdata, 2013) is an HTML specification used to nest semantics within existing content on web pages. Microdata vocabularies provide the semantics, or meaning of an Item. Web developers can design a custom vocabulary or use vocabularies available on the web as shown in Figure 3.

Schema.org (Schema.org, 2013) is a collaborative data or meta-data standardization effort which provides a collection of schemas, i.e., html tags, that webmasters can use to markup their pages in ways recognized by major search providers. Search engines including Bing, Google, Yahoo! and Yandex rely on this markup to improve the display of search results, making it easier for people to find the right web pages. In essence Schema.org is a shared collection of schemas that multiple parties can use and that results in uniformity of meta-data markup by different vendors.

A collection of commonly used markup vocabularies are provided by *Schema.org* schemas which include: Person, Event, Organization, Product, Review, Review-aggregate, Breadcrumb, Offer, Offer-aggregate.

```
<section itemscope itemtype="http://schema.org/Person">
        Hello, my name is
        <span itemprop="name">John Doe</span>,
        I am a
        <span itemprop="jobTitle">graduate research assistant</span>
        at the
        <span itemprop="affiliation">University of Dreams</span>.
        My friends call me
        <span itemprop="additionalName">Johnny</span>.
        You can visit my homepage at
        <a href="http://www.JohnnyD.com" itemprop="url">www.JohnnyD.com</a>.
        <section itemprop="address" itemscope itemtype="http://schema.org/PostalAddress">
                I live at
                <span itemprop="streetAddress">1234 Peach Drive</span>,
                <span itemprop="addressLocality">Warner Robins</span>,
                <span itemprop="addressRegion">Georgia</span>.
        </section>
</section>
```

Figure 3 Microdata Semantic Markup

*RDFa* is a language for meta-data markup. It provides a set of attributes that can be used to carry metadata in an XML language, and has seen very prolific practical use in recent years. *RichSnippets* (Knight, 2013) is a Google technology that allows us to provide detailed snippets i.e., the few lines of information with a search result, along with the product information. Microdata or RDFa formalisms can be used to provide Rich Snippets.

## SA Framework with Semantic Technologies

The work described in (Thomas et al, 2009) delves into an architecture for Semantic Advertising. They advocate a model based on lightweight embedded semantics. The proposed approach is to use some lightweight semantics on a Web page, and RDF to describe advertisements. Then, Semantic-Web technologies for query processing and semantic reasoning can be employed to match Web pages with ads. A semantic reasoning infrastructure for semantic advertising – called TrOWL is proposed.

In this architecture content publishers create the Web page which is embedded with the Semantic-Web data. *RDFa* (RDFa, 2013) and other *Microformats* (Microdata, 2013) are employed for data markup. The next step is to *subscribe* the Website to the Semantic Advertisement system. The publisher is then given a code snippet to include on all pages, at the position where the ad should appear.

The *Advertising Broker* provides a repository for storing the descriptions of web pages, and the descriptions of advertisements and the constraints of the advertisers. The semantics embedded on a page is converted into RDF graphs, and the constraints given by the advertisers will be rewritten as SPARQL (Hitzler, Krotzsch and Rudolph, 2009) queries. By running each query against the repository of graphs extracted from content, we can produce a map of the best advertising for each webpage. When a user requests an ad for a particular page, the system consults the map of appropriate ads and selects the most lucrative one.

## 6. The Importance of Domain Knowledge and Ontologies

True SA enablement technology encompasses many areas of prior and current AI research. This includes work in semantic modeling, user intent understanding, user psychographic profiling, contextual reasoning, natural language processing and knowledge representation and ontologies. It is beyond the scope of this paper to cover all or many of these areas in detail, however we would like to emphasize the importance and prevalence of domain knowledge and ontologies in SA.

Context matching techniques such as (Broder et al., 2007) rely significantly on pre-assembled taxonomies as domain knowledge. Techniques on Wikipedia based matching (Wu et al., 2012) rely on the domain knowledge in Wikipedia for better matching of ads to pages. Ontology efforts have further transcended single organizations with prominent initiative for shared ontologies in the domain.

**GoodRelations** (GoodRelations, 2008) is basically a language that can be used to describe a product offering in precise terms. It can be perceived as a *data dictionary*, a schema or as an ontology. We can use GoodRelations to create a small data package that describes products and their features and prices, stores and opening hours, payment options etc. A major strength of GoodRelations is that it is a standard, *one* vocabulary for many companies and the markup is honored by both traditional search engines such as Google and Yahoo and major retailers such as BestBuy and Target.

The **Knowledge Graph** (Singhal 2012) is a knowledge base used by Google to enhance its search engine's search results with semantic-search information gathered from a wide variety of sources. It provides structured and detailed information about the topic in addition to a list of links to other sites. According to Google, the information in the Knowledge Graph is derived from many sources, including the CIA World Factbook, Freebase, and Wikipedia. As of 2012, its semantic network contained over 570 million objects and more than 18 billion facts about and relationships between different objects that are used to understand the meaning of the keywords entered for the search. The **Graph API** (Graph API, 2013) is the core of Facebook Platform, enabling developers to read from and write data into Facebook. The Graph API presents a simple, consistent view of the Facebook social graph, uniformly representing objects in the graph (e.g., people, photos, events, and pages) and the connections between them (e.g., friend relationships, shared content, and photo tags).

The importance of shared vocabularies or ontologies is being validated from multiple perspectives. Ontologies such as the Knowledge Graph form the core of applications such as Semantic Search. Further, ontologies such as GoodRelations are forming the very basis for successful Semantic E Commerce data exchange and discovery capabilities today.

## 7. Conclusions

We have provided an introduction to Semantic Advertising, our goals of the ideal ad with Semantic Advertising and then a survey of related approaches and technology in this area. With this and our experience in developing and bringing Semantic Advertising technology to the market we conclude with the following perspective:

**Semantic Advertising is the Future of Online Advertising:** We believe that Semantic Advertising (SA) technology will become the standard mechanism for generating effective, revenue-boosting online advertising campaigns. The use of Semantic Advertising (SA) technology will enable the generation of actionable insights from large volumes of noisy data and further apply it to a larger scale. Semantic Advertising (SA) technology will therefore facilitate the individualization of ad messages at scale while allowing marketers to think strategically and enable ad creators to focus on the creative aspects of ad messaging.  We truly believe that the tactic of "mass advertising" is near its life-end and that semantic technologies will facilitate relevant, precise and diverse messaging at scale.

**Shared Vocabularies and Ontologies** Ontologies or taxonomies are a fundamental component to realizing Semantic Advertising. They are a critical component of Semantic Search technology and are core components of today's systems such as Google, Bing and Facebook Graph Search.

**Markup and Standards in Semantic E-Commerce** We have seen significant success in recent years in the area of Semantic E-Commerce with successful initiatives for data markup and shared meta-data. Microdata and RDFa have seen a significantly more prolific adoption rate as opposed to earlier Semantic-web technologies, GoodRelations integrated now with Schema.org has been adopted by most major E-Commerce vendors. Standards for meta-data and data sharing are critical to success in this domain. The present efforts with shared ontologies such as GoodRelations and meta-data standards such as Schema.org must only be further expanded.

## References

Weide, K. 2013. Worldwide and U.S. Internet Ad Spend Report, 4Q12 and FY12. IDC

Tim Berners-Lee, James Hendler and Ora Lassila.2001. The Semantic Web. *Scientific American*, May 2001

Grimes, S. 2010. Breakthrough Analysis: Two + Nine Types of Semantic Search. *InformationWeek*.  Jan (2010).

John, T., 2012. What is Semantic Search? *Techulator*. March, 2012.

Eisenberg, B. 2011. The Conversion Trinity: The 3 Step Magic Formula to Increase Click Throughs & Conversions.  Web:  http://www.bryaneisenberg.com/the-conversion-trinity-the-3-step-magic-formula-to-increase-click-throughs-conversions

Guha, R., McCool, Rob, and Miller, Eric. 2003. Semantic Search.  *WWW Conference* 2003.

Wei Li, Xuerui Wang, Ruofei Zhang, Ying Cui, Jianchang Mao, and Rong Jin. 2010. Exploitation and exploration in a performance based contextual advertising system. In *Proceedings of the 16th ACM*

*SIGKDD international conference on Knowledge discovery and data mining (KDD '10).* ACM, New York, NY, USA, 27-36.

Hammer, M. and McLeod, D. 1990. Database Description with SDM: A Semantic Database Model. *Research Foundations in Object-Oriented and Semantic Database Systems*

Yi Zhang, Arun C. Surendran, John C. Platt, Mukund Narasimhan. 2008. Learning from multi- topic web documents for contextual advertisement. *Knowledge Discovery and Data Mining 2008*: 1051-1059

Kaifu Zhang and Zsolt Katona. 2012. Contextual Advertising. *Marketing Science* 31(6), 980-994. DOI=10.1287/mksc.1120.0740

Jacek Kopecký and Elena Simperl. 2008. Semantic web service offer discovery for e-commerce. In Proceedings of the *10th international conference on Electronic commerce* (ICEC '08). ACM, New York, NY, USA, , Article 29 , 6 pages.

Edward Thomas, Jeff Z. Pan, Stuart Taylor, Yuan Ren, Nophadol Jekjantuk, and Yuting Zhao. 2009. Semantic advertising for web 3.0. In *Proceedings of the Second Future internet conference on Future internet (FIS'2009)*, Tanja Zseby, Reijo Savola, and Marco Pistore (Eds.). Springer-Verlag, Berlin, Heidelberg, 96-105.

Andreas Ekelhart, Stefan Fenz, A. Min Tjoa, and Edgar R. Weippl. 2007. Security issues for the use of semantic web in e-commerce. In Proceedings of the *10th international conference on Business information systems (BIS'07)*, Witold Abramowicz (Ed.). Springer-Verlag, Berlin, Heidelberg, 1-13.

Chen-Yuan Chen, Bih-Yaw Shih, Zih-Siang Chen and Tsung-Hao Chen. 2011. The exploration of internet marketing strategy by search engine optimization: A critical review and comparison. *African Journal of Business Management* Vol. 5(12), pp. 4644-4649, 18 June, 2011

Ciaramita, Massimiliano, Murdock, Vanessa, and Plachouras, Vassilis. 2009. Semantic Associations for Contextual Advertising . *Journal of Electronic Commerce Research*, Vol. 9, No. 1

Andrei Broder. 2008. Computational advertising. In *Proceedings of the nineteenth annual ACM-SIAM symposium on Discrete algorithms (SODA '08).* Society for Industrial and Applied Mathematics, Philadelphia, PA, USA, 992-992.

Andrei Broder, Marcus Fontoura, Vanja Josifovski, and Lance Riedel. 2007. A semantic approach to contextual advertising. In Proceedings of the *30th annual international ACM SIGIR conference on Research and development in information retrieval (SIGIR '07).* ACM, New York, NY, USA, 559-566. DOI=10.1145/1277741.1277837

Roberto Mirizzi, Azzurra Ragone, Tommaso Di Noia, and Eugenio Di Sciascio. 2010. Semantic tags generation and retrieval for online advertising. In Proceedings of the 19th ACM international conference on Information and knowledge management (CIKM '10). ACM, New York, NY, USA, 1089-1098. DOI=10.1145/1871437.1871576

Yannopoulos, Peter. 2011. Impact of the Internet on Marketing Strategy Formulation. *International Journal of Business & Social Science*;2011, Vol. 2 Issue 18, p1

Freebase. 2013. Web: http://www.freebase.com

Singhal, A. 2012. Google Knowledge Graph. Web: http://www.google.com/insidesearch/features/search/knowledge.html

Microdata. 2013. Web: http://www.whatwg.org/specs/web-apps/current-work/multipage/microdata.html#microdata

RDFa. 2013. Rich Structured Data Markup for Web Documents. Web: http://www.w3.org/TR/xhtml-rdfa-primer/

Rajan Varadarajan and Manjit S. Yadav. 2009. Marketing Strategy in an Internet-Enabled Environment: A Retrospective on the First Ten Years of JIM and a Prospective on the Next Ten Years. *Journal of Interactive Marketing*. Volume 23, Issue 1, February 2009, Pages 11–22

Aris Anagnostopoulos, Andrei Z. Broder, Evgeniy Gabrilovich, Vanja Josifovski, and Lance Riedel. 2011. Web Page Summarization for Just-in-Time Contextual Advertising. *ACM Transactions on Intelligent Systems Technology*. 3, 1, Article 14 (October 2011), 32 pages.

Do Viet Phuong and Tu Minh Phuong. 2012. A keyword-topic model for contextual advertising. In Proceedings of the *Third Symposium on Information and Communication Technology (SoICT '12).* ACM, New York, NY, USA, 63-70. DOI=10.1145/2350716.2350728

ZongDa Wu, GuanDong Xu, YanChun Zhang, Peter Dolog, and ChengLang Lu. 2012. An Improved Contextual Advertising Matching Approach based on Wikipedia Knowledge. *Computer Journal*. 55, 3 (March 2012), 277-292. DOI=10.1093/comjnl/bxr056

Hitzler, P., Krotzsch, M. and Rudolph, S. 2009. Foundations of Semantic Web Technologies. *Chapman & Hall/CRC Textbooks in Computing*. Chapman and Hall.

GoodRelations. 2008. The Web Vocabulary for Ecommerce. Web: http://www.heppnetz.de/projects/goodrelations/

Graph API. 2013. Web: https://developers.facebook.com/docs/reference/api/

Schema.org. 2013. Web: http://schema.org/

Knight, L. 2013. The Power of Google Rich Snippets for Ecommerce SEO. Web: http://blog.hubspot.com/power-google-rich-snippets-ecommerce-seo-ht